\documentclass[a4paper,fleqn]{cas-dc}

\usepackage[numbers]{natbib}

\usepackage{amsmath}

\usepackage{xcolor}

\usepackage{graphicx}
\usepackage{amssymb}
\usepackage{booktabs}
\usepackage{float}

\usepackage[capitalize]{cleveref}
\crefname{section}{Sec.}{Secs.}
\Crefname{section}{Section}{Sections}
\Crefname{table}{Table}{Tables}
\crefname{table}{Tab.}{Tabs.}

\def\tsc#1{\csdef{#1}{\textsc{\lowercase{#1}}\xspace}}
\tsc{WGM}
\tsc{QE}

\begin{document}
\let\printorcid\relax
\let\WriteBookmarks\relax
\def\floatpagepagefraction{1}
\def\textpagefraction{.001}

\shorttitle{MFDS-DETR}    

\shortauthors{}  

\title [mode = title]{Accurate Leukocyte Detection Based on Deformable-DETR and Multi-Level Feature Fusion for Aiding Diagnosis of Blood Diseases}



%

\author[1]{Yifei Chen}
\author[1]{Chenyan Zhang}
\author[1]{Ben Chen}
\author[1]{Yiyu Huang}
\author[1]{Yifei Sun}

\author[2]{Changmiao Wang}
\author[3]{Xianjun Fu}
\author[4]{Yuxing Dai}
\author[5]{Feiwei Qin}
\cormark[1]
\author[5]{Yong Peng}
\author[6]{Yu Gao}


\affiliation[1]{organization={HDU-ITMO Joint Institute},
           addressline={Hangzhou Dianzi University}, 
           city={Hangzhou},
           postcode={310018}, 
           country={China}}
\affiliation[2]{organization={Medical Big Data Lab},
           addressline={Shenzhen Research Institute of Big Data}, 
           city={Shenzhen},
           postcode={518172}, 
           country={China}}
\affiliation[3]{organization={School of Artificial Intelligence},
            addressline={Zhejiang College of Security Technology}, 
            city={Wenzhou},
            postcode={325016}, 
            country={China}}
\affiliation[4]{organization={National-local Joint Engineering Research Center for Digitalized Electrical Design Technology},
            addressline={Wenzhou University},
            city={Wenzhou},
            postcode={325035},
            country={China}}
\affiliation[5]{organization={School of Computer Science and Technology},
           addressline={Hangzhou Dianzi University}, 
           city={Hangzhou},
           postcode={310018}, 
           country={China}}
\affiliation[6]{organization={Department of Hematology},
           addressline={Zhejiang Hospital}, 
           city={Hangzhou},
           postcode={310018}, 
           country={China}}
           
\cortext[1]{
Corresponding author.\\
\indent \indent E-mail address: qinfeiwei@hdu.edu.cn (F. Qin).
}



\begin{abstract}
In standard hospital blood tests, the traditional process requires doctors to manually isolate leukocytes from microscopic images of patients' blood using microscopes. These isolated leukocytes are then categorized via automatic leukocyte classifiers to determine the proportion and volume of different types of leukocytes present in the blood samples, aiding disease diagnosis. This methodology is not only time-consuming and labor-intensive, but it also has a high propensity for errors due to factors such as image quality and environmental conditions, which could potentially lead to incorrect subsequent classifications and misdiagnosis. Contemporary leukocyte detection methods exhibit limitations in dealing with images with fewer leukocyte features and the disparity in scale among different leukocytes, leading to unsatisfactory results in most instances. To address these issues, this paper proposes an innovative method of leukocyte detection: the Multi-level Feature Fusion and Deformable Self-attention DETR (MFDS-DETR). To tackle the issue of leukocyte scale disparity, we designed the High-level Screening-feature Fusion Pyramid (HS-FPN), enabling multi-level fusion. This model uses high-level features as weights to filter low-level feature information via a channel attention module and then merges the screened information with the high-level features, thus enhancing the model's feature expression capability. Further, we address the issue of leukocyte feature scarcity by incorporating a multi-scale deformable self-attention module in the encoder and using the self-attention and cross-deformable attention mechanisms in the decoder, which aids in the extraction of the global features of the leukocyte feature maps. The effectiveness, superiority, and generalizability of the proposed MFDS-DETR method are confirmed through comparisons with other cutting-edge leukocyte detection models using the private WBCDD, public LISC and BCCD datasets. Our source code and private WBCCD dataset are available at \href{https://github.com/JustlfC03/MFDS-DETR}{https://github.com/JustlfC03/MFDS-DETR}.
\end{abstract}




\maketitle

\section{Introduction}
The global incidence of severe diseases such as acute leukemia has shown a significant increase in recent years. A primary diagnostic tool for these diseases is the routine blood test, wherein physicians need to examine microscopic images of a patient's blood smear using a microscope. The diagnosis is derived from the proportion and number of different types of leukocytes, or white blood cells. Automated leukocyte classification is commonly employed as a hematological analysis technique to categorize leukocytes in blood images. This technique typically discriminates among various leukocyte subpopulations, including lymphocytes, neutrophils, and monocytes, by scrutinizing attributes such as morphology, size, pigmentation, and nucleolar features to accurately classify distinct leukocyte types \cite{c38bakhri2018analisis}. However, the application of leukocyte classification models often necessitates experienced physicians to manually isolate leukocytes from the microscopic images of a patient's blood, a labor-intensive and time-consuming process prone to errors. In addition, factors such as image quality and environmental conditions can influence the process, leading to potential errors in subsequent classification. These inaccuracies can misdirect the physician's judgment, leading to patient safety concerns. To address these issues, researchers have been exploring leukocyte target detection \cite{C39shitong2006new}. This study aims to automatically and accurately pinpoint the location of leukocytes in blood microscopic images and count the different leukocyte types. This method could expedite the diagnostic and therapeutic decision-making process of doctors, thus enhancing patient care, and has very important research significance.

Traditional leukocyte target detection in blood microscopic images frequently encounters the following challenges: 
\begin{enumerate}[a)]
    \item Different hospitals, using distinct equipment for capturing blood images, generate images with varying color profiles. This variation can lead to diminished effectiveness in leukocyte detection.
    \item The limited number of discernible features in leukocyte images also presents obstacles to efficient detection \cite{c40liu2018path}.
    \item Varying magnification levels across different hospital instruments result in inconsistencies in leukocyte sizes across blood images. Furthermore, inherent size discrepancies among different leukocyte types compound these scale gaps, negatively impacting the efficacy.
    \item Medical microscopy images of leukocytes, unlike natural images, often possess low resolution and divergent imaging modalities. The substantial geometric appearance differences between the targets in these images and objects in natural images pose significant challenges for traditional target detection algorithms \cite{c41woo2018cbam}.
\end{enumerate}

In order to address the challenges associated with leukocyte target detection in blood microscopic imaging, this paper introduces a method based on Multi-level Feature Fusion and Deformable Self-attention DETR (MFDS-DETR). A High-level Screening-feature Fusion Pyramid (HS-FPN) has been designed to facilitate multilevel fusion, taking into account the unique characteristics of leukocytes and the scale gaps observed among different leukocytes. In the HS-FPN, high-level features serve as weights to filter low-level feature information via the channel-attention module. The filtered information is then merged with the high-level features, thereby enhancing the model's feature expression capability. Moreover, to address the problem of feature scarcity in leukocytes, a multi-scale deformable self-attention mechanism is incorporated into the encoder. This assists in extracting the global features of the leukocyte feature map. Subsequently, using self-attention and cross-deformable attention mechanisms, the decoder learns the object to be detected from the global features of the encoder. It then matches the output with the ground truth in a bipartite graph to obtain the location and category of the target. This process enables the automatic detection of leukocytes. 

Compared with existing leukocyte detection methods, the MFDS-DETR effectively addresses the challenge of limited leukocyte features in microscopic blood images. Additionally, it diminishes the impact that scale disparity among various leukocytes in these images has on the efficacy of the modeling process. The primary contributions of this study can be encapsulated as follows:
\begin{itemize}
    \item Within the domain of fine-grained leukocyte detection, we introduce a novel method known as MFDS-DETR. This method, based on multi-scale fusion and deformable self-attention, is made up of four critical components: the backbone network, HS-FPN module, an encoder, and a decoder.
    \item Under the guidance of our team's associated medical practitioners, we labeled the existing publicly available leukocyte classification dataset, LISC, with target frames. We also collaborated with our partner hospitals to develop our own leukocyte detection dataset, WBCCD. The labeled LISC dataset and the WBCCD dataset will be made accessible to other researchers through a download link in our GitHub repository.
    \item The development of the field of white blood cell detection relies heavily on the own situation of the dataset. The existing publicly available LISC and BCCD white blood cell datasets have been collected for a long time, with insufficient size and poor quality. Therefore, we have decided to contribute our WBCCD dataset to other researchers and make important contributions to the development of the field.
    \item We propose the innovative HS-FPN feature fusion module. In contrast to traditional feature fusion methods designed based on natural images, this module has been developed considering the scale gaps inherent in leukocytes. This significant shift greatly enhances the model's feature expression capacity in the leukocyte detection dataset.
    \item Our proposed model, MFDS-DETR, outperforms other advanced and baseline models for leukocyte detection. This is evidenced by the superior detection results obtained on two public datasets, LISC and BCCD, as well as our privately held WBCCD leukocyte fine-grained detection dataset. These results underscore the effectiveness and broad applicability of our model.
\end{itemize}

\section{Related Work}
The YOLO series model, a highly efficient and precise single-stage target detection mechanism, is widely employed in the field of target detection. As such, it forms an integral part of numerous studies focused on leukocyte target detection \cite{c32abas2022yolo}. Wang et al. \cite{c1wang2019deep} utilized the SSD and YOLOv3 models for automated leukocyte detection, achieving the detection of 11 types of peripheral leukocytes. Notably, leukocyte images represent a minor segment of blood images. To address the relatively poor performance of current detection methods when dealing with smaller objects, Wang et al. \cite{c2wang2018so} proposed SO-YOLO, which initially deploys Convolutional Neural Networks (CNNs) to extract image features and subsequently uses YOLO for leukocyte target detection. For further enhancement of the model's performance, Han et al. \cite{c3han2023one} proposed MID-YOLO, a single-stage CNN detector for leukocyte images. This model leverages the attention mechanism and exhibits superior detection performance on the publicly available Raabin-WBC dataset. 
Xu et al. \cite{c4xu2022te}  employed EfficientNet as a backbone network to augment the model's efficiency and flexibility, and proposed the TE-YOLOF detector. 
Khandekar et al. \cite{c33KHANDEKAR2021102690} applied this problem of automated matricellular detection for the diagnosis of acute lymphoblastic leukemia by means of the YOLOv4 target detection algorithm, which is currently used as one of the important auxiliary tools during prescreening. Wang et al. \cite{c5wang2022improved} proposed YOLOv5-CHE, a leukocyte detection model based on an improved YOLOv5. This model addresses the lack of leukocyte samples and inter-class differences, and enhances the model's feature extraction capacity by incorporating a coordinate attention mechanism into the convolutional layer. Considering that the use of a single model may contribute to bias, Xia et al. \cite{c6xia2020ai} designed an integrated model based on YOLOv3, YOLOv3-SPP, and YOLOv3-tiny, achieving an Average Precision (AP) of 88.6\% when the Intersection over Union (IoU) was 0.5. Liu et al. \cite{c7liu2023leukocyte} proposed a leukocyte detection method based on Twin-Fusion-Feature CenterNet (TFF-CenterNet) to mitigate the problems associated with significant variations in leukocyte staining degrees. This approach enhances feature extraction capacity by refining the feature fusion pyramid, thereby addressing the issue of staining degree differences. Notwithstanding the speed of the single-stage target detection model, its detection accuracy still lags behind that of the two-stage target detection model. Kutlu et al. \cite{c8kutlu2020white} employed Faster R-CNN for target detection and experimentally demonstrated that using ResNet-50 as the backbone network yields higher recognition accuracy. Geng et al. \cite{c9Geng2021WhiteBC} improved the Mask R-CNN model's multi-scale feature fusion capability by adding an attention mechanism to the feature fusion pyramid module, consequently boosting the detection accuracy. Polejowska et al. \cite{c34polejowska2023impact} used YOLOv5 in conjunction with the RetinaNet model to accurately quantify lymphocytes by their spatial arrangement and phenotypic characteristics, and validated the performance of the network by applying image modifications such as blurriness, sharpness, brightness, and contrast, etc. Nugraha et al. \cite{c35nugraha2023white} used YOLOv8 in conjunction with DETR for the detection of thousands of leukocytes, and processed multiple subjects in a single image by DETR to improve the detection accuracy. 

However, these studies on leukocyte target detection employ Convolutional Neural Networks (CNNs) to extract features, and then proceed to localize and classify the targets. Such an approach is influenced by the convolutional operator, which falls short of learning the global features of leukocyte images, thereby impeding the accurate localization and classification of peripheral blood leukocytes. Furthermore, the detection efficacy of leukocytes is significantly constrained by the following two challenges:
\begin{enumerate}[a)]
    \item Contrary to the imaging techniques of natural images, leukocyte medical microscopy images are of low resolution. In conjunction with the inherent characteristics of leukocytes, this leads to often insufficient features in the leukocyte images.
    \item The magnification of microscopy instruments varies across different hospitals, and the size of leukocytes is not uniform, resulting in a scale gap among leukocytes.
\end{enumerate}

To address the aforementioned challenges, multi-scale feature fusion is generally applied. This process involves fusing deep features with shallow features, thereby imbuing the shallow features with robust semantic information. Two types of multi-scale feature fusion approaches are prevalent: parallel multi-branch networks and serial jump connection structures. Parallel multi-branch networks typically use different convolutions to extract features from the same feature, subsequently employing splicing to fuse the extracted features. This idea is evident in the Inception module of GoogLeNet \cite{c21szegedy2015going}, which uses various convolutions to extract features from the same feature map and subsequently combines them in the channel dimensions. Similarly, SSP-Net \cite{c22he2015spatial} pools the same feature map in three distinct ways before splicing them to obtain a multi-scale fused feature map. DeepLabv3+ \cite{c23chen2018encoder} adopts the ASSP structure for feature fusion, obtaining features at different scales via hollow convolution and performing scale unification operations by up-sampling. TridentNet \cite{c24li2019scale} and Big-Little Net \cite{c25chen2018big} follow similar approaches, with the latter using the BL module to handle different scales of information more flexibly. In contrast to the parallel splicing method, the serial jump connection structure typically targets the outputs of different layers in the backbone network for multi-scale fusion. Feature Pyramid Networks (FPN \cite{c17lin2017feature}) realizes multi-scale feature fusion by up-sampling the high-level features at a uniform scale and subsequently summing them with the bottom-layer features. However, since FPN is ambiguous for high-level target information, PANet \cite{c19liu2018path} employs bi-directional feature fusion, adding a bottom-up feature fusion module on top of FPN to enhance localization information. Building upon these methods, BiFPN \cite{c18tan2020efficientdet} proposes a more streamlined bi-directional feature fusion, and Balanced FPN \cite{c29pang2019libra} integrates and refines features at all scales before summing them with the original scale features. CE-FPN \cite{c30luo2022fpn} improves the integration and refinement process by utilizing high-level semantic features and the attention mechanism for selective feature fusion. FaPN \cite{c20huang2021fapn} also designs feature selection and feature alignment modules to improve fusion accuracy in response to the potential feature misalignment in FPN.

While these multi-scale feature fusion approaches hold significant reference value, they are fundamentally designed based on natural images. Although some of these methods are effective in leukocyte detection, they do not account for the actual characteristics of leukocyte microscopic images, thereby limiting the detection efficacy of the model. Our proposed MFDS-DETR network model effectively addresses these limitations. The model initially obtains the multi-scale feature map of leukocyte micro-images through the backbone network, then employs the designed HS-FPN for feature fusion. It incorporates a multi-scale deformable self-attention mechanism into the encoder operation to acquire the global features of leukocyte micro-images, and finally uses a decoder to obtain leukocyte location and class. Our innovative model uses high-level features as weights to filter the low-level features, fusing the filtered features with the high-level features to significantly enhance model detection efficacy. Furthermore, by incorporating a multi-scale deformable self-attention mechanism to extract image features, our model markedly reduces complexity and improves detection efficacy.

\section{Method}
\subsection{Overall Architecture}
\begin{figure*}
\centering
\includegraphics[width=0.90\textwidth]{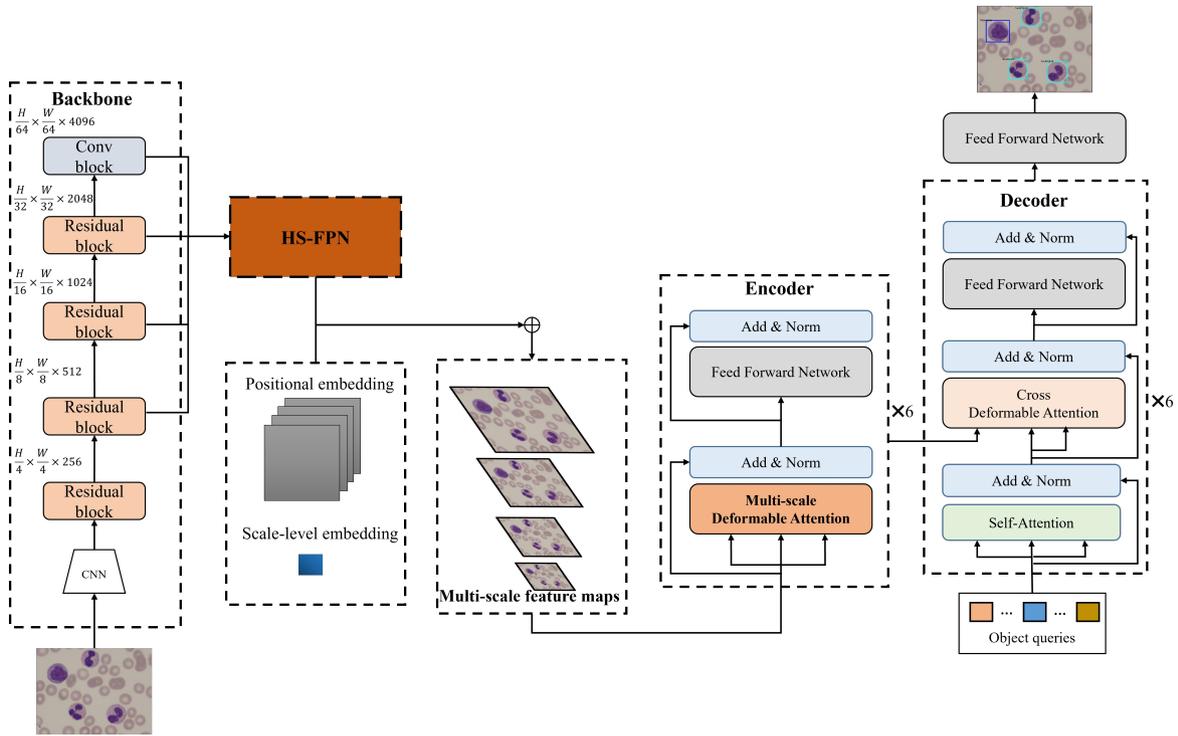}
\caption{The overall architecture of MFDS-DETR comprises four parts: Backbone, High-level Screening-feature Pyramid Networks, Encoder and Decoder.} \label{fig1}
\end{figure*}

The overall structure of the MFDS-DETR model, as depicted in Fig. \ref{fig1}, comprises four key components: the backbone network, HS-FPN, encoder, and decoder. The backbone network primarily serves to extract multi-scale image features of leukocytes, thereby facilitating enhanced feature fusion in subsequent processes. HS-FPN, a feature fusion pyramid designed and improved to accommodate the characteristics of leukocytes, addresses the challenges posed by limited features in leukocyte images and disparities in leukocyte diameters. This is achieved by employing the Channel Attention (CA) module in HS-FPN to leverage high-level semantic features as weights for filtering low-level features. These filtered features are then added point-by-point with the high-level semantic features, thus realizing the multi-scale feature fusion. This ultimately improves the model's feature expression capability. The encoder module's primary function is to learn the global characteristics of leukocyte images. By incorporating a multi-scale deformable self-attention module, the model is equipped to learn the global features of leukocyte images across differing scales. Conversely, the decoder performs bipartite graph matching between the output and ground truth to ascertain the position and category of the target. This is accomplished by learning the objects to be identified from the encoder's global features using self-attention and cross-deformable mechanics.

\subsection{Backbone Network}
An enhanced version of ResNet-50 serves as the backbone network for MFDS-DETR's feature extraction process. ResNet-50 utilizes residual connections to mitigate the vanishing gradient issue, which facilitates convergence and addresses the degradation problem often associated with deep neural networks. Given the paucity of features in leukocyte images, we have augmented our backbone network by adding a convolution block to the original ResNet-50 model. This block is designed to extract deeper semantic information, thereby improving the model's detection effectiveness. In a manner similar to ResNet-50, this convolution block initially reduces the number of channels using a $1\times1$ convolution, subsequently shrinks the feature map size via a $3\times3$ convolution, and, finally, increases the number of channels with another $1\times1$ convolution.

\subsection{High-level Screening-feature Pyramid Networks}

In the leukocyte dataset, the task of leukocyte recognition is challenged by a multi-scale issue, which complicates the model's ability to accurately identify leukocytes. This complexity arises because there are typically differences in diameter among various types of leukocytes, and even identical leukocytes can appear to differ in size when imaged under different microscopes.

\begin{figure*}
\centering
\includegraphics[width=0.90\textwidth]{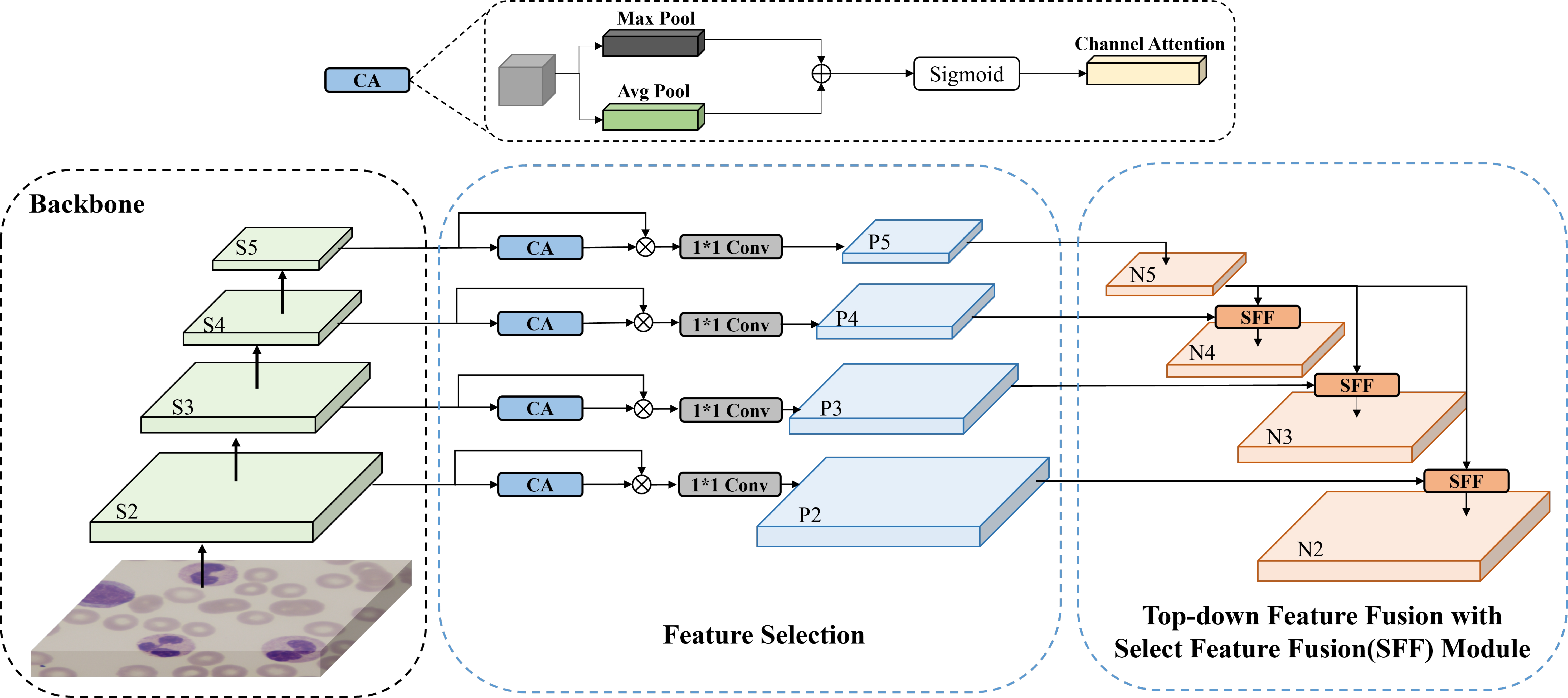}
\caption{The Framework of High-Level Screening-feature Fusion Pyramid Networks comprises two parts: Feature Selection Module and Feature Fusion Module.} \label{fig2}
\end{figure*}

To address the multi-scale challenge inherent in leukocyte datasets, we have developed the Hierarchical Scale-based Feature Pyramid Network (HS-FPN) to accomplish multi-scale feature fusion. This enables the model to capture a more comprehensive array of leukocyte feature information. The structure of HS-FPN is illustrated in Figure \ref{fig2}. The HS-FPN consists of two primary components: (1) the feature selection module. (2) the feature fusion module. Initially, feature maps across different scales undergo a screening process in the feature selection module. Subsequently, high-level and low-level information within these feature maps are synergistically integrated through the Selective Feature Fusion (SFF) mechanism. This fusion yields features with enriched semantic content, which is instrumental in detecting subtle characteristics in leukocyte microscopic images, thereby enhancing the model's detection capabilities. Further elaboration on this SFF mechanism and its impact on model performance will be provided in the subsequent ablation study section.

\textbf{Feature Selection Module: }The CA module and the Dimensional Matching (DM) module play essential roles in this process. The CA module initially processes the input feature map $f_{in}\in R^{C\times H\times W}$, where $C$ represents the number of channels, $H$ denotes the height of the feature map, and $W$ signifies the width of the feature map. This feature map undergoes two pooling layers—global average pooling and global maximum pooling—after which, the resulting features are combined. Subsequently, the Sigmoid activation function is utilized to determine the weight value of each channel, culminating in the weight of each channel, $f_{CA}\in R^{C\times1\times1}$. Pooling serves several fundamental purposes: it down-samples and reduces the dimensionality of the feature map; eliminates redundant data, compresses features, and decreases the parameter count; and achieves translation, rotation, and scale invariance. The CA module employs global average pooling and global maximum pooling to calculate the average and maximum values of each channel, respectively. The primary aim of maximum pooling is to extract the most pertinent data from each channel, while average pooling is designed to uniformly acquire all data from the feature map, minimizing excessive loss. Therefore, the combination of these pooling methods in the CA module facilitates the extraction of the most representative information from each channel, while concurrently minimizing information loss. The filtered feature map is subsequently generated by multiplying the weight information with the feature map at the corresponding scale. Prior to the feature fusion, dimensionally matching the feature maps across various scales is crucial as they possess different numbers of channels. To accomplish this, the DM module applies a $1\times1$ convolution to reduce the number of channels for each scale feature map to 256.

\begin{figure*}
\centering
\includegraphics[width=0.90\textwidth]{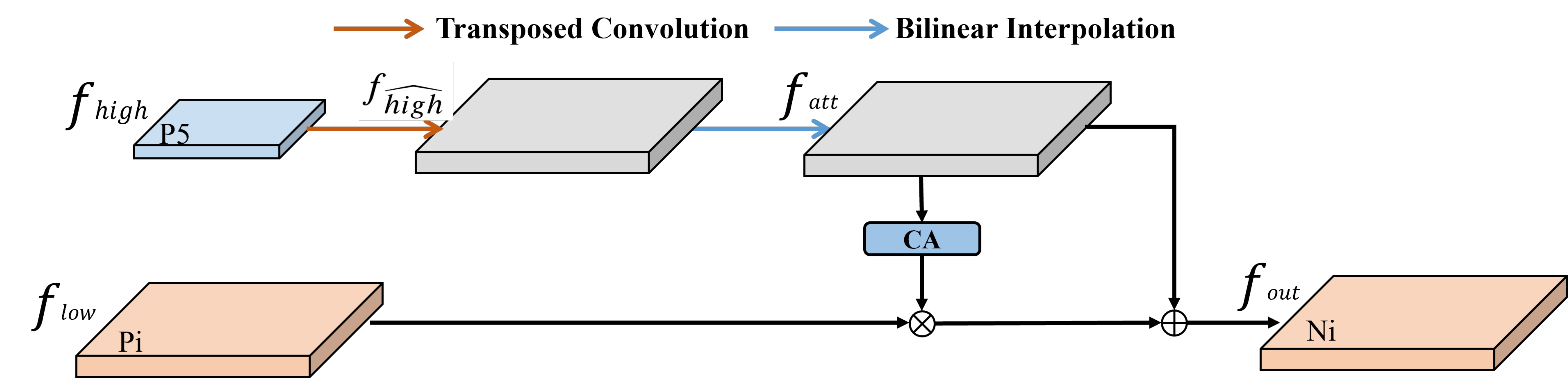}
\caption{The Framework of SFF Module. The combination of transposed convolution and bilinear interpolation is used to process high-level features and achieve purposeful feature fusion.} \label{fig3}
\end{figure*}

\textbf{Feature Fusion Module: }The multi-scale feature maps generated by the backbone network present high-level features with an abundance of semantic information but relatively coarse target localization. Conversely, low-level features offer precise target locations but contain limited semantic information. A common solution to this conundrum involves the direct pixel-wise summation of the up-sampled high-level features and low-level features, enriching each layer with semantic information. However, this technique does not perform a feature selection but simply sums the pixel values across multiple feature layers. To address this limitation, in this study, we have developed the SFF module. This module strategically fuses features by using high-level features as weights to filter essential semantic information embedded within the low-level features. As depicted in Fig. \ref{fig3}, given an input high-level feature $f_{high}\in R^{C\times H\times W}$ and an input low-level feature $f_{low}\in R^{C\times H_1\times W_1}$, the high-level feature is initially expanded using a transposed convolution (T-Conv) with a stride size of 2 and a convolution kernel of $3\times3$, resulting in a feature size of $f_{\widehat{high}}\in R^{C\times2H\times2W}$. Afterwards, to unify the dimensions of the high-level features and the low-level features, we use Bilinear Interpolation to either up-sample or down-sample the high-level features to obtain $f_{att}\in R^{C\times H_1\times W_1}$. The CA module is then employed to convert the high-level features into corresponding attention weights to filter the low-level features, upon obtaining features with consistent dimensions. Finally, the filtered low-level features are fused with the high-level features to enhance the model's feature representation and obtain $f_{out}\in R^{C\times H_1\times W_1}$. Equations (\ref{eq1}) and (\ref{eq2}) illustrate the fusion process for feature selection:

\begin{equation} \label{eq1} f_{att}=BL\left(T-Conv\left(f_{high}\right)\right), \end{equation}

\begin{equation} \label{eq2} f_{out}=f_{low}\ast CA\left(f_{att}\right)+f_{att}. \end{equation}

In the image sampling process, we utilize a combination of transposed convolution and bilinear interpolation to restore the high-level features' scale. Bilinear interpolation is simple and fast, achieving direct manipulation of the pixels for image zooming. The advantages of transposed convolution include: 1) adapting to the data through learnable parameters, such that the output not only enlarges the feature map but also reconstructs the input in the form of convolution, which is achieved through a convolution kernel realizing convolution operations after the feature map expansion by filling in zeros; and 2) it can handle non-uniform sampling problems by sampling different regions of the input image at different output image locations. Our ablation experiments further substantiate that the combination of transposed convolution and bilinear interpolation outperforms the use of only bilinear interpolation.

\subsection{Deformable Self-attention Module}
The Deformable Self-Attentive Module primarily comprises two components: the Offset Module and the Attention Module. A comprehensive description of their respective implementations ensues.
\begin{figure*}
\centering
\includegraphics[width=0.90\textwidth]{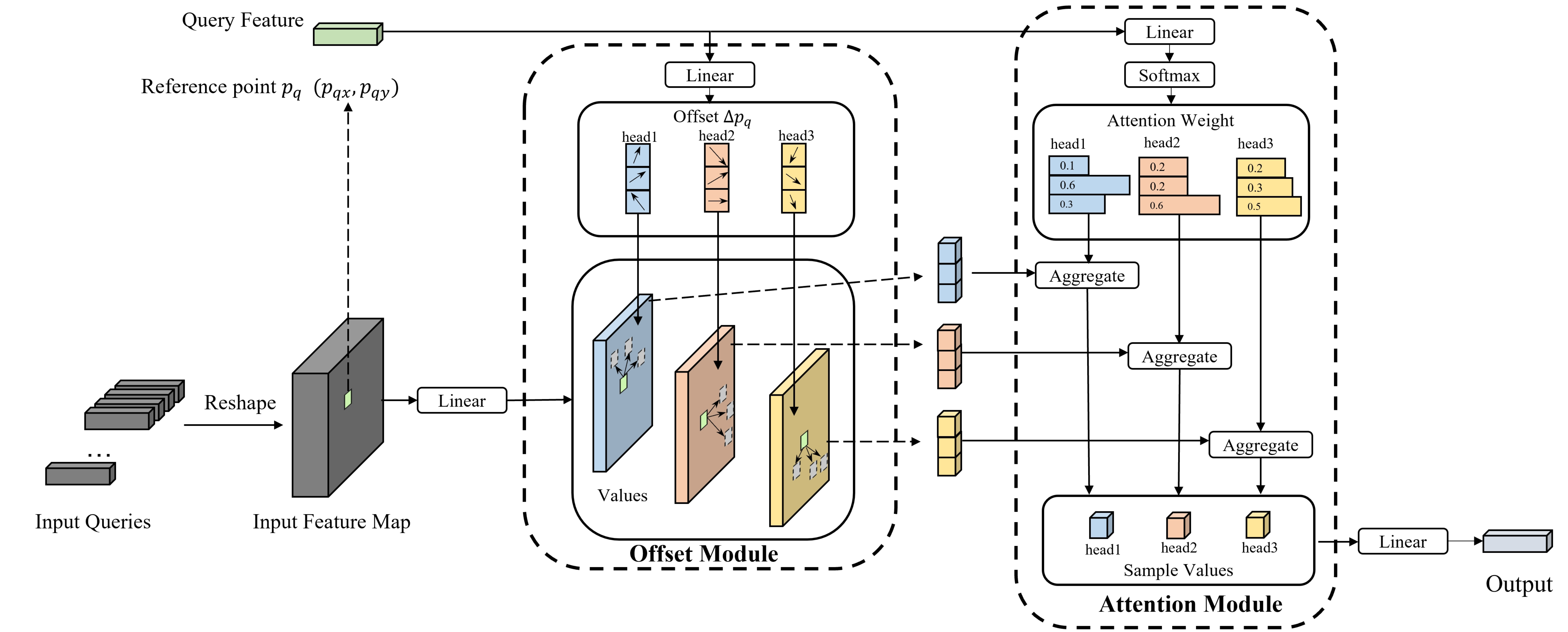}
\caption{The Framewrok of Deformable Self-attention Module comprises two parts: Offset Module and Attention Module.} \label{fig4}
\end{figure*}

\textbf{Offset Module: }As depicted in Fig. \ref{fig4}, the vector must be transformed into a feature map prior to its integration into the Offset Module, and then the input query vector is generated, taking into account the coordinates of the reference points. A linear transformation is applied to the query vector to yield the offset $\Delta p_q$, while a similar approach is applied to the input feature map to derive the content feature map. Subsequently, the point of interest, or the sampling point, for each reference point is determined based on the reference points' offsets. Bilinear interpolation is then employed to achieve the output $offset_{value}$ for each point. As represented in Fig. \ref{fig4}, each query vector possesses $H$ attention heads, while each attention head is associated with $K$ offset points. In the experiments conducted within this study, values for $H$ and $K$ were established at 8 and 4, respectively.

\textbf{Attention Module: }As depicted in Fig. \ref{fig4}, the process within the Attention Module initiates with a linear transformation of the input query vector. Following this, the Softmax function is applied to generate the weight vector for each offset. The output of each offset, as determined in the Offset Module, is then multiplied by its corresponding weight vector. The subsequent results are aggregated to yield $\textit{Sample}_{\textit{value}}$. Subsequently, the attention heads corresponding to each reference point are concatenated, yielding the final vector, denoted as $\textit{Sample}_{\textit{output}}$. Lastly, a linear transformation is conducted on the sampled output vector, resulting in the ultimate output value. The equations for this process are elaborated in (\ref{eq3}), (\ref{eq4}), (\ref{eq5}), and (\ref{eq6}):
\begin{equation} \label{eq3}
    \textit{Weight}=\textit{Softmax}(WQ),
\end{equation}
\begin{equation} \label{eq4}
    \textit{Sample}_{\textit{value}}=\sum\limits_{k=1}^{K}\textit{offset}_{\textit{value}}\ast \textit{Weight},
\end{equation}
\begin{equation} \label{eq5}
    \begin{aligned}
    \textit{Sample}_{\textit{output}}=&\textit{Concat}(\textit{Sample}_{\textit{value}}^1,\\
    & \textit{Sample}_{\textit{value}}^2, \cdots, \textit{Sample}_{\textit{value}}^H),
    \end{aligned}
\end{equation}
\begin{equation} \label{eq6}
    \textit{Output}=W\ast \textit{Sample}_{\textit{output}}.
\end{equation}
Given that MFDS-DETR employs the HS-FPN module to facilitate multi-scale feature fusion for the backbone network's feature inputs, the encoder inputs comprise multi-scale feature maps. To extract the feature information of leukocytes at various scales, a multi-level deformable attention module was utilized. This module is not restricted to enabling offset learning for the scale of the input reference points; it also promotes offset learning from different scales based on the relative positions of the normalized reference points at these scales. The module computes the offset output vector for each scale. Subsequently, the resulting vectors from different scales are weighted and integrated to yield the final output vector. The equation utilized to calculate the multi-scale deformable attention is shown in (\ref{eq7}):
\begin{equation} \label{eq7}
    \begin{aligned}
    &\textit{MSDAM}= \\
    &\sum\limits_{h=1}^{H}\left (\sum\limits_{l=1}^{L}\sum\limits_{k=1}^{K}\textit{Weight}_{hlqk}\ast V(p_q+\Delta p_{hlqk})\right ),
    \end{aligned}
\end{equation}
where $H$ represents the number of attention heads, $L$ signifies the number of multi-scales, and $K$ denotes the number of sampling points. The weight of the reference point $q$ at the $h$-th head and the $k$-th sampling point at the $L$-layer scale is expressed as $\textit{Weight}_{hlqk}$. $V$ corresponds to the feature map that has been transformed from the input vector. The absolute coordinates of the reference point, within the range [0,1], are represented by $p_q$. The offset of the reference point $q$ at the $h$-th head and the $k$-th sampling point of the $L$-layer scale is denoted as $\Delta p_{hlqk}$. The sampling value that corresponds to the current sampling point in the content feature map is given by $V(p_q+\Delta p_{hlqk})$.

\subsection{Encoder and Decoder}
The encoder plays a critical role in the extraction of global features from leukocyte images. The input to the encoder is a multi-level feature map, integrating both spatial location encoding and scale encoding, as demonstrated in Fig. \ref{fig1}. Each layer within the encoder is composed of a deformable self-attention module and a Feed Forward Network (FFN). Given the significant influence of the reference point position in deformable self-attention, it is initialized through the determination of the center coordinates of each scale pixel point and subsequent normalization of the width and height of the various scale feature maps. Once the reference point positions are defined, the multi-level deformable self-attention module generates output vectors. In this study, the deformable self-attention was configured to eight attention heads, with each head focusing on offsets in different directions. To mitigate the risk of gradient vanishing and expedite model convergence, the output vectors are produced and normalized using a residual structure in the Add and Norm method. These output vectors are then processed through the FFN network structure - a multilayer perceptron responsible for expanding and reducing dimensionality, which enables the model to learn more nonlinear correlations between features. Ablation tests conducted within this study demonstrated optimal performance with a six-layer encoder in the MFDS-DETR model.

The decoder serves a pivotal function in establishing the relationships between various detected feature representations and in identifying the precise location and class of the target. As illustrated in Fig. \ref{fig1}, each layer of the decoder is comprised of two components: the self-attention feature extraction module and the cross-attention feature extraction module. The self-attention feature extraction module consists of a self-attention module and an FFN, which are structurally analogous to those in the encoder. However, unlike the encoder, the keys and queries of the cross-attention feature extraction module are derived from the output location encoding (Object Queries), while its values are sourced from the global features extracted from the final layer of the encoder.

\subsection{Joint Loss Function}
The composite loss function is presented in Equation (\ref{eq10}): 

\begin{equation} 
\label{eq10}     
\mathcal{L}_{\textit{final}}=\sum\limits_{i=1}^{N}(\mathcal{L}_{class}+\mathcal{L}_{box}), 
\end{equation} 

which is composed of three principal components: classification loss, regression loss, and auxiliary loss. The classification and regression loss functions are employed to optimize the model and ascertain the most optimal matching value. Conversely, the auxiliary loss is used to expedite the model's convergence. Primarily, the auxiliary loss calculates the classification loss and regression loss for each output layer of the decoder.

\subsubsection{Classified Losses}
As the decoder encompasses 100 object query boxes, and the genuine target of a leukocyte image typically comprises only 2 or 3 leukocytes, a significant imbalance between positive and negative samples can occur. Therefore, we incorporated the Focal loss function to address this imbalance. The computation of the Focal loss function is presented in Equation (\ref{eq11}):
\begin{equation} \label{eq11}
    \mathcal{L}_{\textit{focal}(y,\hat y)}=-\sum\limits_{i=1}^{C}\alpha_iy_{im}(1-prob_{im})^\gamma \log(prob_{im}).
\end{equation}
The Focal loss function incorporates two hyperparameters: $\alpha$ and $\gamma$. Here, $\alpha$ signifies the proportion of each category within the dataset, while $\gamma$ is assigned a value of 2. In this setting, $\alpha$ is employed in the Focal loss function to counterbalance the uneven proportions in the dataset. Meanwhile, $\gamma$ symbolizes the weighted contribution of both difficult and straightforward samples to the total loss function.

\subsubsection{Regression Loss}
As the L1 loss function is influenced by the size of the leukocyte input image, we addressed this issue by formulating a novel regression loss function. This was achieved by combining the L1 loss function with the GIoU loss function, as illustrated in Equation (\ref{eq12}):
\begin{equation} \label{eq12}
\begin{aligned}
    \mathcal{L}_{box}(box_y,box_{\hat y})=&\lambda_{GIoU}\mathcal{L}_{GIoU}(box_y, box_{\hat y}) \\
    &+\lambda_{\mathrm{L1}} |box_y-box_{\hat{y}}|,
    \end{aligned}
\end{equation}
where $box_y$ and $box_{\hat{y}}$ denote the prediction box and target box paired using the Hungarian algorithm, and $\lambda_{GIoU}$ and $\lambda_{L1}$ are two hyperparameters that represent the weights of the GIoU and L1 losses, and $\mathcal{L}_{GIoU}$ denotes the GIoU loss function, and the L1 loss function is the absolute value of the difference between the target box and the prediction box positions.

\subsubsection{Ancillary Losses}
Contrasting with the two previously mentioned loss functions, the auxiliary loss is primarily applied to expedite model training. While the original model exclusively utilizes the output from the final encoder layer to predict the target, the auxiliary loss capitalizes on each layer of the encoder output for prediction. Further, the classification and loss functions are computed for each layer to streamline model training. As a result, the ultimate loss function is calculated as shown in Equation (\ref{eq10}), where each decoder layer functions as the model's terminal layer for prediction. Here, 'N' represents the count of decoder layers.

\section{Experiment}
\subsection{Dataset}
\begin{figure*}
\centering
\includegraphics[width=0.80\textwidth]{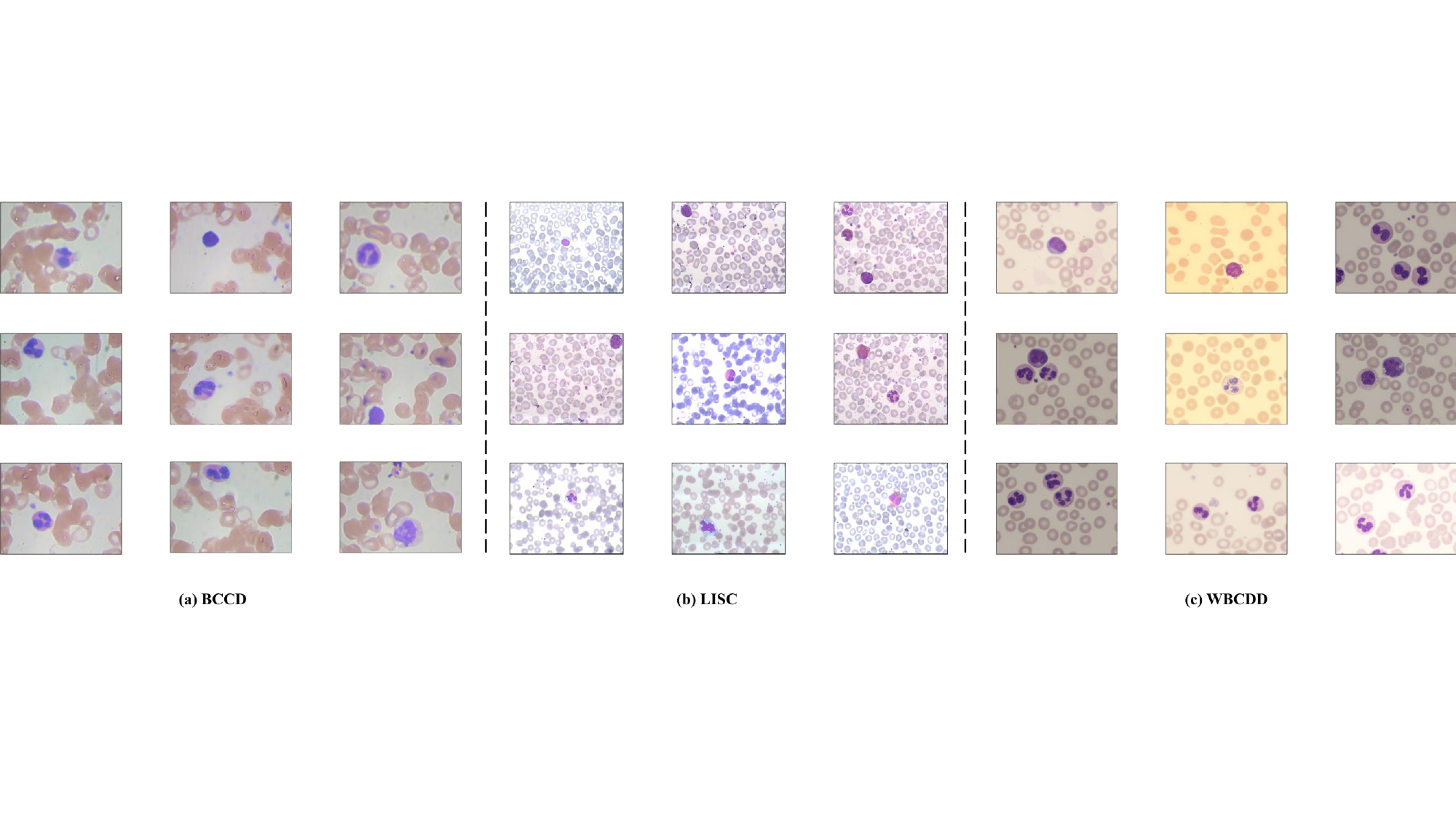}
\caption{Presentation of blood images in various datasets.} \label{fig5}
\end{figure*}

For validation, we employ a total of three datasets: the Leukocyte Detection Dataset (WBCDD), LISC \cite{c10rezatofighi2011automatic}, and BCCD \cite{c11Vatathanavaro2018WhiteBC}. The publicly accessible LISC and BCCD datasets were used to evaluate the model's generalizability. In contrast, the WBCDD dataset was specifically assembled for this research. Fig. \ref{fig5} displays leukocyte images from each dataset, while Table \ref{tab1} and Table \ref{tab2} detail the counts of various cell types present in each dataset.

We sourced the WBCDD dataset from multiple local hospitals. In laboratory settings, doctors scrutinized patients' blood cell images under a microscope to compile the images for this dataset. Qualified clinicians guided the annotation of leukocyte bounding boxes using the LabelMe software. This dataset includes a total of 684 samples, encompassing 1,257 leukocytes of five different types: neutrophils (NEU), eosinophils (EOS), monocytes (MON), basophils (BAS), and lymphocytes (LYM). Prior to the experiment, we partitioned the dataset, with the training set consisting of 540 image samples and the test set comprising 144 samples. 

The LISC dataset, an early leukocyte dataset, is made up of blood images derived from healthy individuals' peripheral blood samples. The slides underwent staining via the Gismo-right method and were examined at 100x magnification using a Sony model SSCD50AP camera and an Axioskope 40 microscope. A specialist hematologist then classified 250 blood images into five granular categories. However, given that this dataset is categorical, we annotated it with white blood cells using the LabelMe tool under medical professionals' guidance to render it suitable for our target detection model. Before the experiment, we divided this dataset into a training set, comprising 200 images, and a testing set, containing 50 images.
\begin{table*}[hp]
\caption{The number of different categories of leukocytes in the LISC and WBCDD datasets.}\label{tab1}
\begin{tabular*}{\tblwidth}{LLLLLLL}
\toprule
Dataset & NEU & MON & EOS & LYM & BAS & Total cells \\
\midrule
LISC \cite{c10rezatofighi2011automatic} & 53 & 51 & 44 & 55 & 57 & 260 \\
WBCDD & 1,008 & 51 & 14 & 171 & 13 & 1,257 \\
\bottomrule
\end{tabular*}
\end{table*}

The Blood Cell Count Dataset (BCCD) is a public dataset of labeled blood cell images. The Gismo-right staining method was used to prepare this dataset, which was then visualized under a standard optical microscope fitted with a 100x magnification CCD color camera. A blood expert provided the annotations for this assemblage. The collection, which includes a total of 364 images, is divided into three categories: leukocytes (WBCs), red blood cells (RBCs), and platelets. Contrasting with the first two datasets, the BCCD dataset includes annotations for both blood cells and platelets. Consequently, the blood cells in this dataset display dense distributions, featuring instances of target adhesion and occlusion among distinct blood cells. Prior to initiating the experiments, the dataset was partitioned into a training set with 292 images and a test set containing 72 images.
\begin{table*}[hp]
\caption{The number of different categories of leukocytes in the BCCD dataset.}\label{tab2}
\begin{tabular}{lllll}
\toprule
Dataset & RBC & WBC & Platelets & Total cells \\
\midrule
BCCD \cite{c11Vatathanavaro2018WhiteBC} & 4,155 & 372 & 361 & 4,888 \\
\bottomrule
\end{tabular}
\end{table*}

The development of white blood cell detection relies heavily on the size and quality of datasets. Due to the fact that the existing publicly available LISC dataset has been collected for too long from now, and there is also a problem of insufficient dataset size, the model cannot be trained well on this dataset. In addition, the annotations in the existing publicly available BCCD dataset not only include blood cells, but also platelets, resulting in a dense distribution of blood cells in the dataset. There is also target adhesion and occlusion between different blood cells, greatly affecting the actual effectiveness of the model. Therefore, we collected data from local hospitals and collaborated with professional laboratory doctors to observe patient blood images through a microscope. We also used the LabelMe tool to label white blood cell target boxes, and ultimately established the WBCCD dataset. At the same time, we have decided to make this dataset publicly available to other researchers in the field to promote further development in the field of white blood cell detection.

\subsection{Experimental setup}
We employed the Python language and implemented the MFDS-DETR model using the PyTorch deep learning framework. Given the slow convergence of the MFDS-DETR model on smaller datasets, we initially trained it on the publicly available MS COCO dataset and then fine-tuned it on the leukocyte target detection dataset utilizing the concept of transfer learning. The experimental setup incorporated an NVIDIA GeForce RTX 3090 with 24 GB of RAM as the hardware configuration, with Ubuntu 20.04 selected as the operating system. 

The model was trained with a batch size of 100. The backbone network's learning rate was configured at 0.00002, while the encoder and decoder had a learning rate of 0.0002. The HS-FPN had a learning rate of 0.0003. We employed the StepLR learning rate decay strategy, which reduces the learning rate to 0.1 times the original value every 40 batches. The AdamW optimizer was used for model tuning, with parameters $\beta_1=0.9$, $\beta_2=0.999$, and a weight decay set at 0.0001.

\subsection{Comparison of other methods}
We assess the effectiveness and generalizability of the MFDS-DETR model by executing a comparative experiment on several distinct leukocyte target detection datasets. This comparative experiment is conducted individually on three datasets: WBCDD, LISC, and BCCD. We juxtapose the MFDS-DETR model with traditional models in the domain of leukocyte target detection, such as Faster R-CNN \cite{c12ren2015faster}, SSD \cite{c13liu2016ssd}, RetinaNet \cite{c14lin2017focal}, DETR \cite{c15carion2020end}, Deformable DETR \cite{c16zhu2020deformable}, TE-YOLOF \cite{c4xu2022te} and YOLOv5-ALT \cite{c37guo2023blood}. 

The outcomes of the MFDS-DETR model on the leukocyte detection dataset are presented in Table \ref{tab3}. Our proposed MFDS-DETR model achieves 79.7\% and 97.2\% for AP and $AP_{50}$, respectively, on the WBCDD dataset. These results demonstrate that our model can effectively enhance the accuracy of leukocyte target detection through the use of multi-scale and global feature extraction. The MFDS-DETR model proves more efficient compared to the traditional two-stage target detection model (Faster R-CNN), with improvements of 21.5\% and 23.5\% in AP and $AP_{50}$, respectively. When contrasted with conventional single-stage object detection models utilizing multi-level feature extraction, such as SSD, the MFDS-DETR model boosts the $AP$ and $AP_{50}$ by 15.5\% and 16.7\%, respectively. Additionally, in comparison to object detection models leveraging global feature extraction like DETR, the MFDS-DETR model enhances the $AP$ and $AP_{50}$ by 12.9\% and 10.8\%, respectively. Furthermore, the calculated AP values for each class of leukocytes in Table \ref{tab3} facilitate deeper analysis of the model's enhancements. In the WBCDD dataset, noticeably, the AP values for eosinophils and lymphocytes show significant improvement compared to the baseline model (Deformable DETR), with increases of 10.5\% and 6.6\%, respectively. Additionally, our model MFDS-DETR stands out among currently cutting-edge Leukocyte detection methods (TE-YOLOF and YOLOv5-ALT).
\begin{table*}[hp]
\caption{Comparison of recognition effectiveness of different detection models on the WBCDD dataset.}\label{tab3}
\begin{tabular}{lllllllll}
\hline
Models & $\text{AP}_{\text{NEU}}$ & $\text{AP}_{\text{MON}}$ & $\text{AP}_{\text{EOS}}$ &
$\text{AP}_{\text{LYM}}$ & $\text{AP}_{\text{BAS}}$ & AP & $\text{AP}_{\text{50}}$ & $\text{AP}_{\text{75}}$ \\
\hline
Faster R-CNN \cite{c12ren2015faster}             & 84.9          & 53.1          & 41.5          & 73.1          & 38.2          & 58.2          & 73.7          & 72.4 \\
SSD \cite{c13liu2016ssd}                      & 83.1          & 48.0          & 49.0          & 67.2          & 73.9          & 64.2          & 80.5          & 77.9 \\
RetinaNet \cite{c14lin2017focal}                & 85.1          & 47.3          & 31.1          & 66.4          & 7.9           & 47.6          & 57.0          & 55.3 \\
DETR \cite{c15carion2020end}                     & 84.1          & 53.4          & 52.4          & 73.6          & 70.5          & 66.8          & 86.4          & 82.5 \\
Deformable DETR \cite{c16zhu2020deformable}          & 84.2          & 68.7          & 74.5          & 73.7          & 73.1          & 74.9          & 94.4          & 93.6 \\
TE-YOLOF \cite{c4xu2022te} & 86.9 & 59.3 & 69.3 & 79.7 & 47.2 & 68.5 & 88.7 & 86.5 \\
YOLOv5-ALT \cite{c37guo2023blood} & 88.2 & 62.7 & 74.2 & 72.2 & 59.4 & 71.3 & 98.2 & 93.4 \\
\textbf{MFDS-DETR(Ours)} & \textbf{87.1} & \textbf{71.5} & \textbf{85.0} & \textbf{80.3} & \textbf{74.9} & \textbf{79.7} & \textbf{97.2} & \textbf{96.8} \\
\hline
\end{tabular}
\end{table*}

\begin{table*}[hp]
\caption{Comparison of recognition effectiveness of different detection models on the LISC dataset.}\label{tab4}
\begin{tabular}{lllllllll}
\hline
Models & $\text{AP}_{\text{NEU}}$ & $\text{AP}_{\text{MON}}$ & $\text{AP}_{\text{EOS}}$ &
$\text{AP}_{\text{LYM}}$ & $\text{AP}_{\text{BAS}}$ & AP & $\text{AP}_{\text{50}}$ & $\text{AP}_{\text{75}}$ \\
\hline
Faster R-CNN \cite{c12ren2015faster}             & 83.3          & 71.4          & 80.2          & 70.2          & 77.5          & 76.5          & 100.0         & 96.9 \\
SSD \cite{c13liu2016ssd}                      & 73.7          & 72.0          & 61.5          & 68.9          & 75.3          & 70.3          & 96.1          & 92.7 \\
RetinaNet \cite{c14lin2017focal}                & 46.1          & 22.0          & 19.7          & 69.8          & 19.7          & 37.0          & 52.1          & 47.6 \\
DETR \cite{c15carion2020end}                     & 82.1          & 76.0          & 80.6          & 72.5          & 77.7          & 77.8          & 98.9          & 98.9 \\
Deformable DETR \cite{c16zhu2020deformable}          & 79.6          & 77.2          & 82.6          & 72.9          & 78.2          & 78.1          & 100.0         & 95.4 \\
TE-YOLOF \cite{c4xu2022te} & 81.2 & 79.2 & 81.1 & 69.3 & 76.2 & 77.4 & 100.0 & 94.9 \\
YOLOv5-ALT \cite{c37guo2023blood} & 84.3 & 74.5 & 84.1 & 77.1 & 59.5 & 75.9 & 98.8 & 97.9 \\
\textbf{MFDS-DETR(Ours)} & \textbf{75.2} & \textbf{81.8} & \textbf{83.9} & \textbf{74.2} & \textbf{82.5} & \textbf{79.5} & \textbf{99.9} & \textbf{98.7} \\
\hline
\end{tabular}
\end{table*}

\begin{table*}[hp]
\caption{Comparison of recognition effectiveness of different detection models on the BCCD dataset.}\label{tab5}
\begin{tabular}{lllllllll}
\hline
Models & $\text{AP}_{\text{RBC}}$ & $\text{AP}_{\text{WBC}}$ & $\text{AP}_{\text{Placelets}}$ & AP & $\text{AP}_{\text{50}}$ & $\text{AP}_{\text{75}}$ \\
\hline
Faster R-CNN \cite{c12ren2015faster}             & 61.0          & 80.9          & 52.4          & 64.8          & 93.9          & 73.4          \\
SSD \cite{c13liu2016ssd}                      & 55.8          & 73.8          & 45.7          & 58.4          & 92.6          & 67.0          \\
RetinaNet \cite{c14lin2017focal}                & 54.0          & 71.7          & 40.5          & 55.4          & 90.1          & 60.0          \\
DETR \cite{c15carion2020end}                     & 63.6          & 81.6          & 46.6          & 64.0          & 95.4          & 70.5          \\
Deformable DETR \cite{c16zhu2020deformable}          & 59.0          & 83.1          & 48.4          & 63.5          & 92.4          & 74.0          \\
TE-YOLOF \cite{c4xu2022te} & 53.3 & 78.7 & 43.2 & 58.4 & 90.6 & 84.1 \\
YOLOv5-ALT \cite{c37guo2023blood} & 74.4 & 82.3 & 55.0 & 70.6 & 97.4 & 73.2 \\
\textbf{MFDS-DETR(Ours)} & \textbf{63.7} & \textbf{84.8} & \textbf{53.6} & \textbf{67.4} & \textbf{95.3} & \textbf{74.1} \\
\hline
\end{tabular}
\end{table*}

To further evaluate the model's generalization capability, we subjected the model to identical experiments on the LISC and BCCD datasets. As shown in Table \ref{tab4} and Table \ref{tab5}, the model displays superior performance in terms of $AP$ and $AP_{75}$ on the LISC dataset. Despite its $AP_{50}$ being marginally lower than both Faster R-CNN and the baseline model (Deformable DETR) by a mere 0.1\%, it achieves optimal detection results for all cell types, with the exception of neutrophils. This can be attributed to neutrophils in the LISC dataset being segmented nucleus cells, characterized by distinct nuclei differing from other types of white blood cells. Moreover, on the BCCD dataset, our model attains optimal results across all indicators compared to other models, thereby further substantiating that our proposed target detection model, MFDS-DETR, can classify and localize leukocytes with greater precision.

\subsection{Ablation study}
The efficacy of leukocyte testing is significantly influenced by both the size and quality of the dataset employed. The LISC dataset, collected some time ago, is marked by its limited size and suboptimal quality. In contrast, the annotations within the BCCD dataset encompass not only leukocytes but also platelets, leading to a densely populated representation of blood cells. This results in frequent instances of target adhesion and occlusion among different blood cell types within the dataset. Given these characteristics, we employed these two public datasets strictly for the purpose of assessing the model's generalizability across varying conditions and opted not to conduct ablation experiments with them. Our exploration of the influence of different components and specific parameter configurations on the performance of the MFDS-DETR model was solely conducted using the WBCDD dataset.

\subsubsection{Comparison of different multi-level feature fusion strategies}
\begin{figure}
\centering
\includegraphics[width=0.5\textwidth]{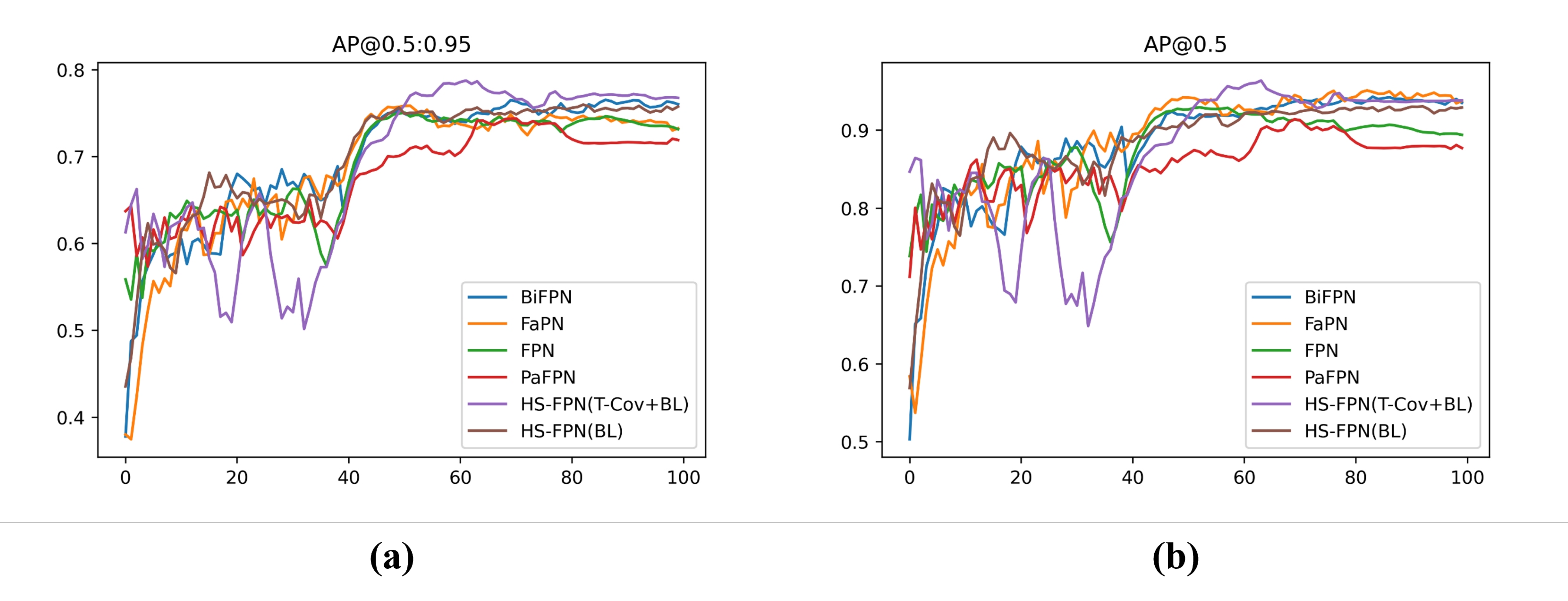}
\caption{Plot of AP values for different multi-scale feature fusion methods.} \label{fig6}
\end{figure}

Given the scale gap among leukocytes, we designed the HS-FPN module to selectively fuse high-level semantic information with low-level features to enable more accurate localization and classification of leukocytes. To demonstrate the effectiveness of the HS-FPN module in fusing multi-scale features, we compared it with other multi-scale feature fusion methods, such as FPN \cite{c17lin2017feature}, BiFPN \cite{c18tan2020efficientdet}, PaFPN \cite{c19liu2018path}, and FaPN \cite{c20huang2021fapn}. As depicted in Table \ref{tab6}, HS-FPN outperforms FPN by improving AP by an additional 3.6\%. Moreover, it enhances $AP_{50}$ and $AP_{75}$ by 4.1\% and 4.6\%, respectively. When compared to other state-of-the-art FPN models, HS-FPN demonstrates superior performance in white blood cell detection. These results affirm that HS-FPN efficiently selects high-level semantic information, using it as weights to filter low-level features, thereby facilitating a more effective fusion of high-level semantic information with low-level attributes in white blood cell images. The AP curves for different FPN variants are displayed in Fig. \ref{fig6}. Fig. \ref{fig6}(a) presents the AP curves, while Fig. \ref{fig6}(b) specifically illustrates the AP curve for an IoU threshold of 0.5.

Table \ref{tab6} further illustrates that, within the HS-FPN module, we are able to enhance the model's detection capabilities more effectively by employing a combination of up-sampling techniques such as transposed convolution and bilinear interpolation, compared to the sole use of bilinear interpolation for up-sampling.
\begin{table}
\caption{Comparison of AP, $\text{AP}_{\text{50}}$, $\text{AP}_{\text{75}}$ for different multiscale feature fusion methods.}\label{tab6}
\begin{tabular}{llll}
\hline
Feature fusion mode & AP & $\text{AP}_{\text{50}}$ & $\text{AP}_{\text{75}}$ \\
\hline
FPN \cite{c17lin2017feature}                      & 76.1          & 93.1          & 92.2          \\
BiFPN \cite{c18tan2020efficientdet}                      & 77.7          & 94.0          & 94.0          \\
PaFPN \cite{c19liu2018path}                      & 77.4          & 93.2          & 92.3          \\
FaPN \cite{c20huang2021fapn}                       & 78.0          & 95.0          & 94.9          \\
HS-FPN(BL)                 & 76.8          & 93.5          & 93.5          \\
\textbf{HS-FPN(T-Conv+BL)} & \textbf{79.7} & \textbf{97.2} & \textbf{96.8} \\
\hline
\end{tabular}
\end{table}

\subsubsection{The comparison between the number of encoder layers and decoder layers}
\begin{figure}
\centering
\includegraphics[width=0.5\textwidth]{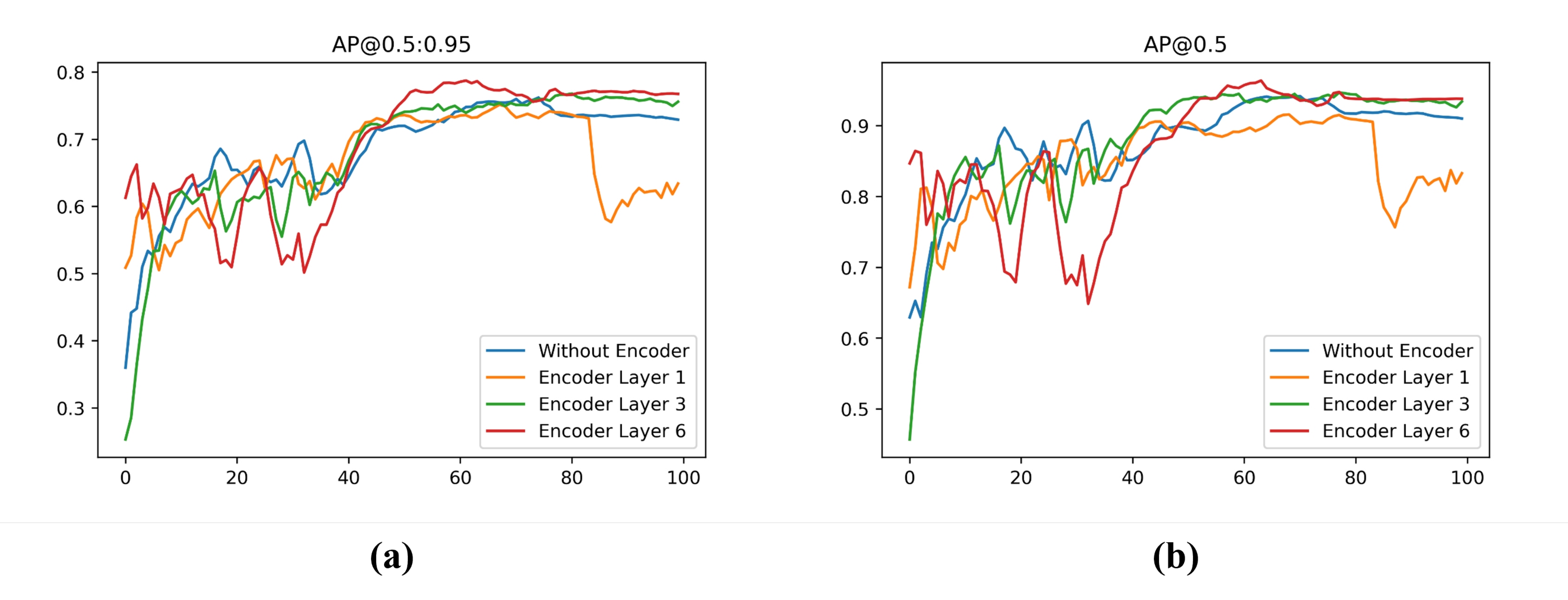}
\caption{Plot of AP values corresponding to different encoder layers.} \label{fig7}
\end{figure}

The role of the encoder in the MFDS-DETR is instrumental in enabling the model to learn global features. To underscore the importance of global feature learning, we examined the impact of varying the number of encoder layers. As depicted in Table \ref{tab7}, the omission of an encoder results in a 2.8\% decrease in AP, with corresponding declines in $AP_{50}$ and $AP_{75}$ by 3.2\% and 3.0\%, respectively. The use of a single encoder layer did not enhance performance and instead caused a slight degradation. This indicates that a solitary encoder layer is inadequate for accurately extracting global features from images. The AP increases in proportion to the number of encoder layers, highlighting the fundamental role of the encoder in processing leukocytes and their spatial configurations. This experiment strongly emphasizes the importance of learning global features. Fig. \ref{fig7} illustrates the AP curves for a range of encoder layer counts. Fig. \ref{fig7}(a) showcases the AP curve, while Fig. \ref{fig7}(b) presents the AP curve for an IoU threshold of 0.5.

\begin{table}
\caption{Comparison of AP, $\text{AP}_{\text{50}}$, $\text{AP}_{\text{75}}$ with different layers of encoders.}\label{tab7}
\begin{tabular}{clll}
\hline
Encoder layers & AP & $\text{AP}_{\text{50}}$ & $\text{AP}_{\text{75}}$ \\
\hline
0          & 76.9          & 94.0          & 93.8          \\
1          & 76.8          & 92.1          & 92.1          \\
3          & 77.9          & 96.0          & 96.0          \\
\textbf{6} & \textbf{79.7} & \textbf{97.2} & \textbf{96.8} \\
\hline
\end{tabular}
\end{table}

\begin{figure}
\centering
\includegraphics[width=0.5\textwidth]{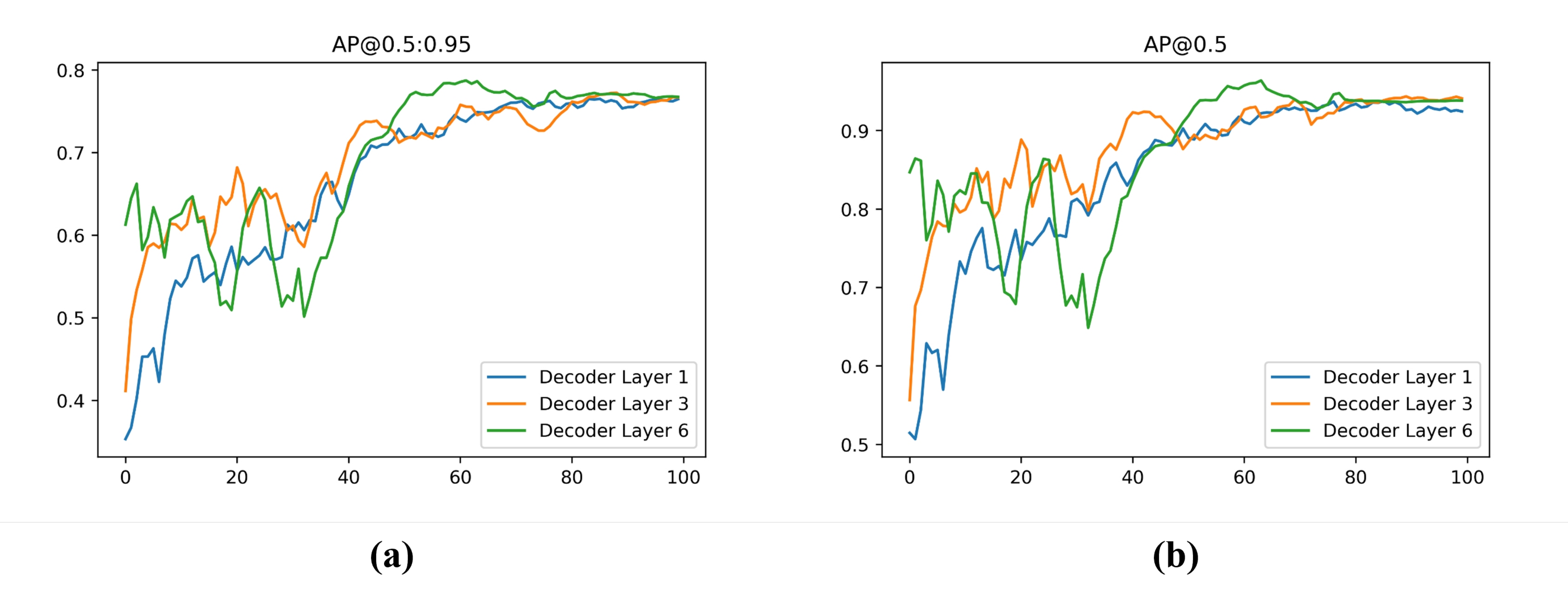}
\caption{Plot of AP values corresponding to different decoder layers.} \label{fig8}
\end{figure}

The MFDS-DETR decoder is instrumental in modelling the interactions among various detection feature representations. In this study, we manipulated the number of decoder layers to substantiate the significance of the decoder. As shown in Table \ref{tab8} and Fig. \ref{fig8}, the detection performance consistently diminished with a reduction in the number of decoder layers. Specifically, when we used a single encoder layer as the model's prediction output, there was a decline in the AP, $AP_{50}$, and $AP_{75}$ by 1.6\%, 2.7\%, and 2.6\%, respectively.

\begin{table}
\caption{Comparison of AP, $\text{AP}_{\text{50}}$, $\text{AP}_{\text{75}}$ with different layers of decoders.}\label{tab8}
\begin{tabular}{clll}
\hline
Encoder layers & AP & $\text{AP}_{\text{50}}$ & $\text{AP}_{\text{75}}$ \\
\hline
1          & 78.1          & 94.5          & 94.2          \\
3          & 78.8          & 95.4          & 95.2          \\
\textbf{6} & \textbf{79.7} & \textbf{97.2} & \textbf{96.8} \\
\hline
\end{tabular}
\end{table}

\subsubsection{Comparison of different position coding methods}
In the MFDS-DETR model, we encounter three distinct types of position encodings: output encoding (object queries), scale encoding, and spatial position encoding. To enhance the learning of an image's global features, the image must undergo a serialization operation. Spatial position encoding is then added to each sequence block to denote the position of the serialized image within the original image. Unlike the DETR model, the spatial position encoding in MFDS-DETR is only added in the encoder because the deformable attention is employed and the reference point in the decoder is determined by the output, negating the need to add key values to the spatial position encoding. However, for multi-scale feature maps, the spatial position encoding will be identical across different scales, leading to an inability to distinguish positions. Hence, to distinguish inputs across different scales, MFDS-DETR introduces scale position encoding in addition to spatial position encoding. Output encoding is an essential positional coding that outputs the location of the prediction frame.

\begin{figure}
\centering
\includegraphics[width=0.5\textwidth]{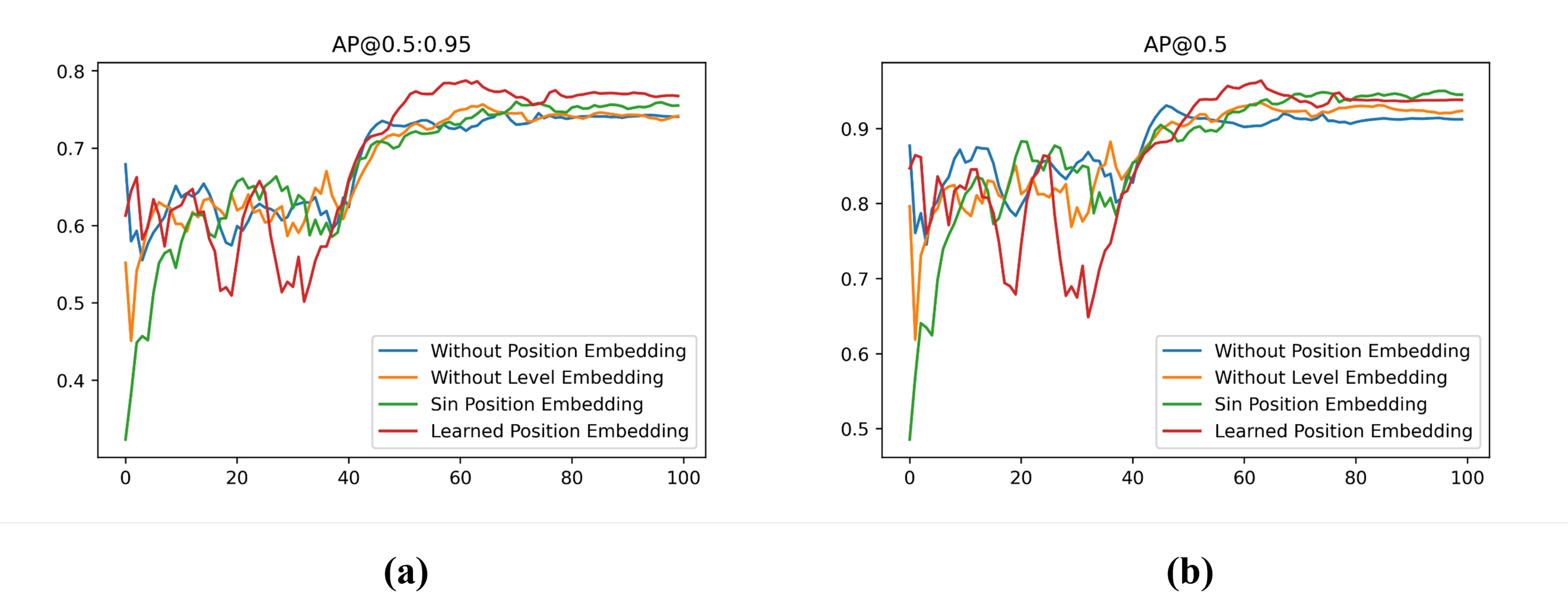}
\caption{Plot of AP values corresponding to different positional codes.} \label{fig9}
\end{figure}

Spatial position encoding primarily utilizes two methods: Learned Position Encoding (Learned PE) and Fixed Encoding (Sin PE). Experimental results ascertain the necessity of both scale encoding and spatial location encoding. As depicted in Table \ref{tab9}, the absence of spatial position encoding and scale encoding leads to a respective decrease in AP by 3.1\% and 3.4\%. Furthermore, fixed position encoding proves more effective in the encoder than learned position encoding. Fig. \ref{fig9} presents the AP curves for various position encodings.

\begin{table*}[hp]
\caption{Comparison of AP, $\text{AP}_{\text{50}}$, $\text{AP}_{\text{75}}$ corresponding to different positional codes.}\label{tab9}
\begin{tabular}{llllll}
\hline
Scale encoding   & Spatial position encoding & Output encoding & AP & $\text{AP}_{\text{50}}$ & $\text{AP}_{\text{75}}$ \\
\hline
Learned          & -                         & Learned                  & 76.6          & 93.0          & 93.0          \\
\textbf{Learned} & \textbf{Sin}              & \textbf{Learned}         & \textbf{79.7} & \textbf{97.2} & \textbf{96.8} \\
Learned          & Learned                   & Learned                  & 77.6          & 94.6          & 94.6          \\
-                & Sin                       & Learned                  & 76.3          & 93.4          & 92.9          \\
\hline
\end{tabular}
\end{table*}

\subsubsection{Comparison of joint loss functions}
To underscore the significance of various components in the model's joint loss function, we conducted ablation experiments using different combinations of loss permutations. In the joint loss function of MFDS-DETR, three types of losses are employed: classification loss, regression loss, and auxiliary loss. The regression loss encapsulates the L1 bounding box loss function and the GIoU loss function. The classification loss function is pivotal to the model's training and cannot be omitted. Hence, in this study, we trained a model excluding the bounding box distance loss and another one excluding the GIoU loss. Furthermore, to establish the importance of the auxiliary loss, we trained a model that does not incorporate this loss. 

\begin{figure}
\centering
\includegraphics[width=0.5\textwidth]{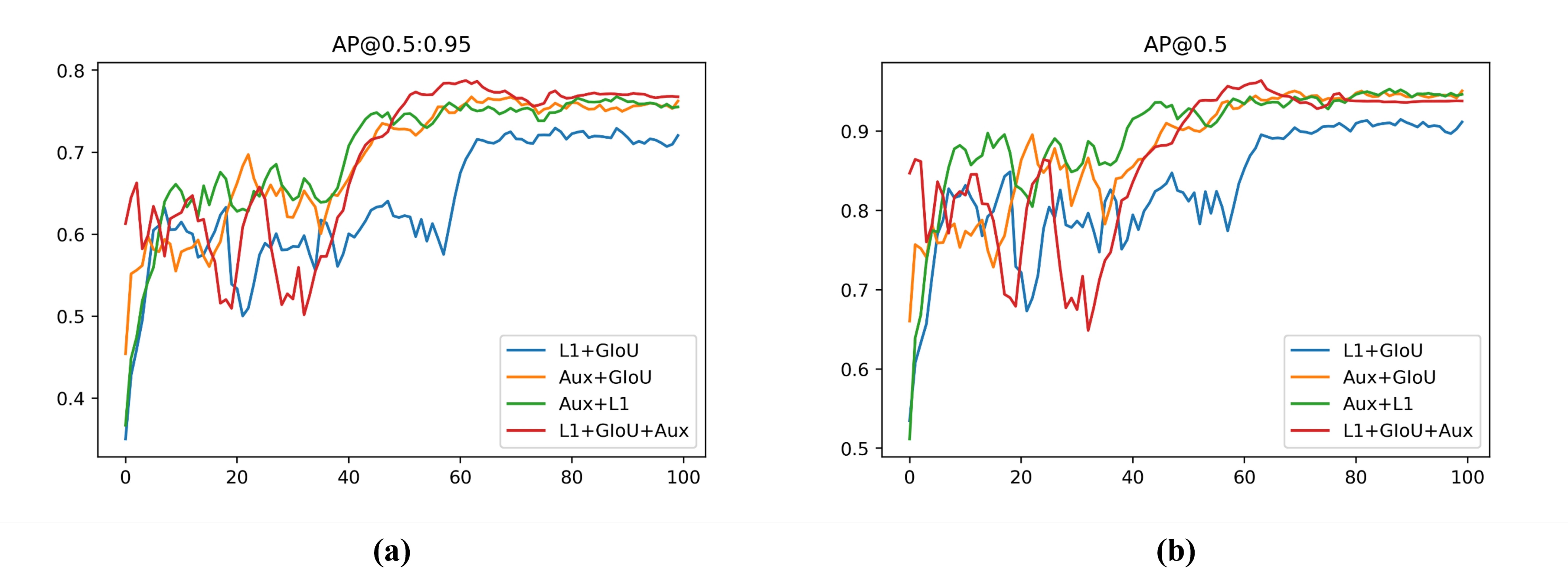}
\caption{Plot of AP values corresponding to different loss functions.} \label{fig10}
\end{figure}

As depicted in Fig. \ref{fig10} and Table \ref{tab10}, the GIoU loss proves to be more crucial than the L1 loss. The absence of the L1 loss results in a 1.4\% decrease in AP, while the omission of the GIoU loss leads to a 1.9\% decrease. Moreover, the experiments demonstrated a significant 4.5\% drop in AP compared to the model trained without the auxiliary loss, solidifying the importance of this loss function in the model's training.

\begin{table*}[hp]
\caption{Comparison of AP, $\text{AP}_{\text{50}}$, $\text{AP}_{\text{75}}$ corresponding to different loss functions.}\label{tab10}
\begin{center}
\begin{tabular}{llllll}
\hline
Classified loss     & L1 loss             & GIoU loss           & Auxiliary loss      & AP & $\text{AP}_{\text{50}}$ \\
\hline
$\checkmark$          & $\checkmark$          & -                     & $\checkmark$          & 77.8          & 95.9          \\
$\checkmark$          & -                     & $\checkmark$          & $\checkmark$          & 78.3          & 95.3          \\
$\checkmark$          & $\checkmark$          & $\checkmark$          & -                     & 75.2          & 92.9          \\
\textbf{$\checkmark$} & \textbf{$\checkmark$} & \textbf{$\checkmark$} & \textbf{$\checkmark$} & \textbf{79.7} & \textbf{97.2} \\
\hline
\end{tabular}
\end{center}
\end{table*}

\subsection{Model visualization analysis}
\begin{figure*}
\centering
\includegraphics[width=0.80\textwidth]{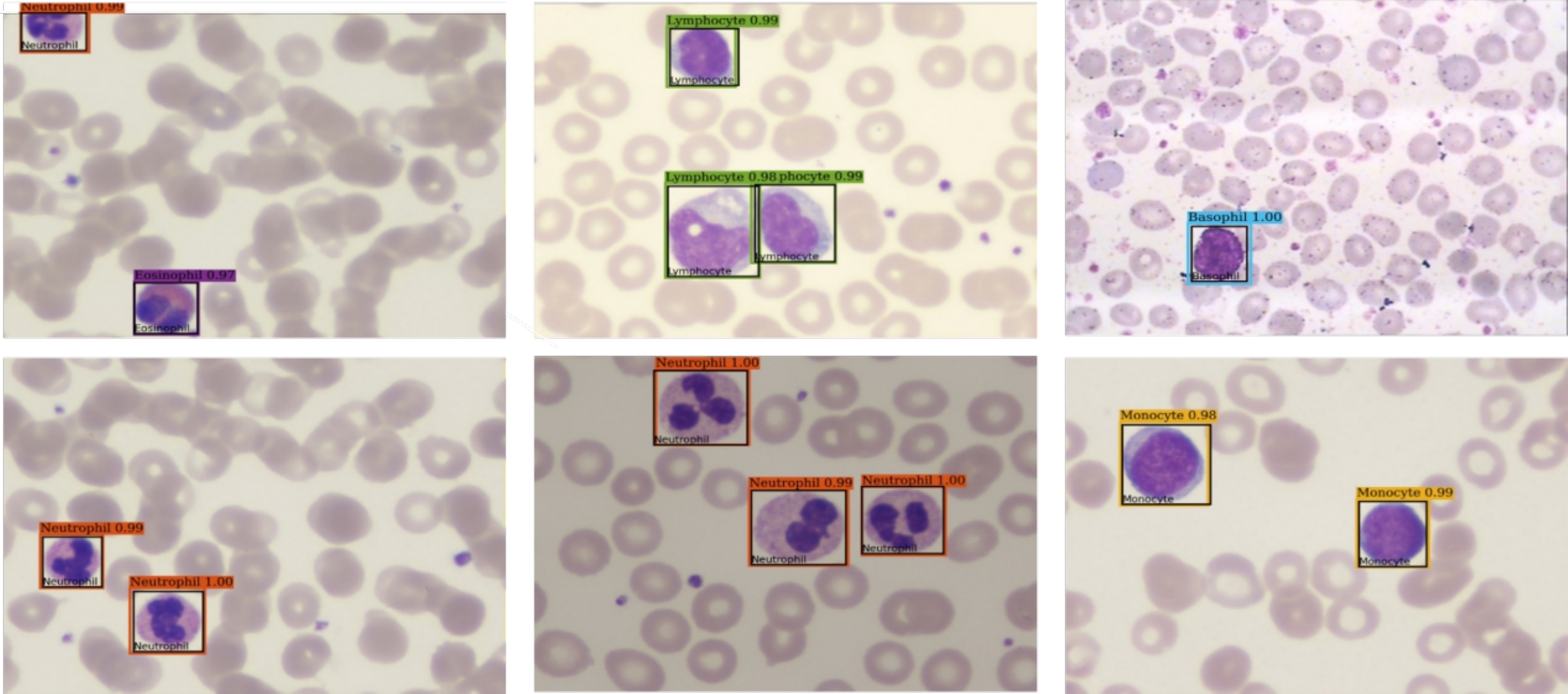}
\caption{Prediction results of MFDS-DETR on the WBCDD dataset (black boxes indicate Ground Truth and different colored boxes indicate its prediction boxes).} \label{fig11}
\end{figure*}

To illustrate the efficacy of the model's predictions more vividly, Fig. \ref{fig11} showcases the model-predicted categories and locations, as well as the original image categories and target boxes from the WBCDD dataset. The Ground Truth is indicated by the black box, while the other boxes represent the results and confidence levels predicted by our MFDS-DETR model. The green, orange, purple, blue, and yellow boxes denote lymphocytes, neutrophils granulocytes, eosinophils, basophils, and monocytes, respectively. As observed from the figure, our model acquires high prediction confidence and precise prediction positioning for all five types of leukocyte images, thus demonstrating its remarkable effectiveness and significant application value.

\section{Conclusion}
In this paper, we introduce the MFDS-DETR model, detailing its network structure and joint loss function. The network structure comprises a backbone network, a HS-FPN module, an encoder, and a decoder. The primary function of the backbone network is to extract multi-scale features from leukocyte images, thus enabling subsequent multi-scale feature fusion. The HS-FPN, tailored specifically for the unique characteristics of leukocytes, employs a channel attention module as a means to use high-level feature maps as weights for screening low-level features. These screened features are subsequently amalgamated with the high-level features, thereby enriching the low-level features with significant semantic information. The encoder leverages deformable self-attention to extract the image's global features, while the decoder employs self-attention and cross-deformable attention for learning the location of the target. 

Moreover, the joint loss function, specifically designed for the model, integrates classification loss, regression loss, and auxiliary loss. The model is optimized through classification and regression losses, with the primary objective to identify the most suitable matching value. The convergence of the model is expedited by the auxiliary loss, thereby computing the classification and regression losses of the decoder outputs at each layer. 

In the subsequent segment on comparison experiments, we juxtapose our specially designed MFDS-DETR model with other advanced leukocyte target detection models using three datasets, namely, WBCDD, LISC, and ALL-IDB, as a testament to our model's effectiveness and generalizability. Furthermore, we employ ablation experiments on the WBCDD dataset to establish the importance of the model's key components, such as position encoding, encoder, decoder, and the joint loss function. Finally, we utilize visualization and model effects analysis to further substantiate the effectiveness of our model.

The advancement of leukocyte assays is significantly constrained by the size and quality of the available datasets. The LISC dataset, a longstanding publicly accessible resource, is limited by its small scale. Furthermore, the BCCD dataset not only annotates various blood cells but also includes platelets, leading to a densely populated and complex dataset characterized by instances of target adhesion, occlusion, and substandard image quality. Recognizing these limitations, we have decided to release the WBCCD dataset to researchers in the domain, with the anticipation that this high-quality dataset will catalyze progress in the field of leukocyte detection.

Our study offers a substantial contribution to the field; however, we must acknowledge its limitations. To enhance the robustness and generalizability of the MFDS-DETR model, future research should focus on amassing larger and more varied datasets for further validation. Moreover, in light of the swift progress in medical imaging technology and deep learning methodologies, ongoing refinement and adaptation of our model are imperative to maintain its relevance and utility in practical applications.

\section*{Acknowledgements}
This work was supported by National Key Research and Development Program of China (No. 2023YFE0114900), Natural Science Foundation of Zhejiang Province (No. LY21F020015), National Natural Science Foundation of China (No. 61972121), the Open Project Program of the State Key Laboratory of CAD\&CG (No. A2304), Zhejiang University, GuangDong Basic and Applied Basic Research Foundation (No. 2022A1515110570), Innovation Teams of Youth Innovation in Science and Technology of High Education Institutions of Shandong Province (No. 2021KJ088) and National College Student Innovation and Entrepreneurship Training Program (No. 202310336074). The authors would like to thank the reviewers for their valuable comments and pertinent suggestions.







\printcredits


\end{document}